\documentclass{article}
\usepackage{arxiv}
\usepackage[utf8]{inputenc} 
\usepackage[T1]{fontenc}    
\usepackage{hyperref}       
\usepackage{url}            
\usepackage{booktabs}       
\usepackage{amsfonts}       
\usepackage{nicefrac}       
\usepackage{microtype}      
\usepackage{lipsum}

\usepackage{cite}
\usepackage{amsmath,amssymb}
\usepackage{algorithmic}
\usepackage{graphicx}
\usepackage{textcomp}
\usepackage{xcolor}
\usepackage{multirow}
\usepackage{subfigure}

\title{Unsupervised Feedforward Feature (UFF) Learning \\ for Point Cloud Classification and Segmentation}

\author{Min Zhang\\ Media Communications Lab\\ University of Southern California\\ Los Angeles, CA, USA\\ \texttt{zhan980@usc.edu}\\ \And Pranav Kadam\\ Media Communications Lab\\ University of Southern California\\ Los Angeles, CA, USA\\ \texttt{pranavka@usc.edu}\\ \And Shan Liu\thanks{This work was supported by Tencent Media Lab.}\\ Tencent Media Lab\\ Tencent America\\ Palo Alto, CA, USA\\ \texttt{shanl@tencent.com}\\ \And C.-C. Jay Kuo\\ Media Communications Lab\\ University of Southern California\\ Los Angeles, CA, USA\\ \texttt{cckuo@sipi.usc.edu}}

\begin{document}
\maketitle
\begin{abstract}
In contrast to supervised backpropagation-based feature learning in deep neural networks (DNNs), an unsupervised feedforward feature (UFF) learning scheme for joint classification and segmentation of 3D point clouds is proposed in this work. The UFF method exploits statistical correlations of points in a point cloud set to learn shape and point features in a one-pass feedforward manner through a cascaded encoder-decoder architecture. It learns global shape features through the encoder and local point features through the concatenated encoder-decoder architecture. The extracted features of an input point cloud are fed to classifiers for shape classification and part segmentation. Experiments are conducted to evaluate the performance of the UFF method. For shape classification, the UFF is superior to existing unsupervised methods and on par with state-of-the-art DNNs. For part segmentation, the UFF outperforms semi-supervised methods and performs slightly worse than DNNs. 
\end{abstract}


\section{Introduction}\label{sec:introduction}

Processing and analysis of 3D Point clouds are challenging since the 3D spatial coordinates of points are irregular so that 3D points cannot be properly ordered to be fed into deep neural networks (DNNs). To deal with the order problem, a certain transformation is needed in the deep learning pipeline. Transformation of a point cloud into another form often leads to information loss. Several DNNs have been designed for point cloud classification and segmentation in recent years. Examples include PointNet \cite{qi2017pointnet}, PointNet++ \cite{qi2017pointnet++}, DGCNN \cite{wang2018dynamic}, and PointCNN \cite{li2018pointcnn}. They address the point order problem and reach impressive performance in tasks such as classification, segmentation, registration, object detection, etc. However, DNNs rely on expensive labeled data. Furthermore, due to the end-to-end optimization, deep features are learned iteratively via backpropagation. To save both labeling and computational costs, it is desired to obtain features in an unsupervised and feedforward one-pass manner. 

Unsupervised or self-supervised feature learning for 3D point clouds was investigated in \cite{achlioptas2017learning, yang2018foldingnet, doersch2017multi, hassani2019unsupervised}. Although no labels are needed, the learned features are not as powerful as the supervised one with degraded performance. Recently, two light-weight point cloud classification methods, PointHop \cite{zhang2020pointhop} and PointHop++ \cite{zhang2020pointhop++}, were proposed. Both of them have an unsupervised feature learning module, and their performance is comparable with that of the PointNet \cite{qi2017pointnet}. 

By generalizing the PointHop, we propose a new solution for joint point cloud classification and part segmentation here. Our main contribution is the development of an unsupervised feedforward feature (UFF) learning system with an encoder-decoder architecture. UFF exploits the statistical correlation between points in a point cloud set to learn shape and point features in a one-pass feedforward manner. It obtains the global shape features with an encoder and the local point features using the encoder-decoder cascade. The shape/point features are then fed into classifiers for shape classification and point classification (i.e. part segmentation). Experiments are conducted to evaluate the performance of the UFF method. For shape classification, UFF is superior to all previous unsupervised methods and on par with state-of-the-art DNNs. For part segmentation, UFF outperforms semi-supervised methods and performs slightly worse than DNNs. The rest of the paper is organized as follows. Related work is given in Sec. \ref{sec:related}. The UFF encoder is detailed in Sec. \ref{sec:method}. Experimental results are shown in Sec. \ref{sec:experiment}. Finally, concluding remarks are given in Sec. \ref{sec:conclusion}. 

\section{Related Work} \label{sec:related}

{\bf DNNs.} Deep learning networks for point clouds have attracted a lot of attention nowadays. State-of-the-art DNNs for point cloud classification and segmentation are variants of the PointNet \cite{qi2017pointnet}. PointNet learns features of each point individually using multi-layer perceptrons (MLPs) and aggregates all point features with a symmetric function to address the order problem. PointNet inspires follow-ups, including PointNet++ \cite{qi2017pointnet++}, DGCNN \cite{wang2018dynamic}, and PointCNN \cite{li2018pointcnn}. They incorporate the information of neighboring points to learn more powerful local features. As to unsupervised learning, autoencoders were proposed in \cite{achlioptas2017learning, yang2018foldingnet} for feature learning. FoldingNet \cite{yang2018foldingnet} trains a graph-based feature encoder and adopts a folding-based decoder to deform a 2D grid to the underlying object surface. 

{\bf Successive Subspace Learning (SSL).} Kuo {\em et al.} \cite{kuo2016understanding, kuo2018data, kuo2019interpretable} introduced an unsupervised feature learning method based on successive subspace approximation. This strategy with other supporting ideas \cite{chen2020pixelhop} is named ``successive subspace learning (SSL)". Recently, PointHop \cite{zhang2020pointhop} was proposed for point cloud classification based on SSL. It conducts local-to-global attribute building for 3D point clouds through an iterative one-hop information exchange process. Attributes for a point are derived from the distribution of points in its 3D neighborhood. The Saab transform \cite{kuo2019interpretable} is used to control the rapid increase in the attribute dimension. As an extension, PointHop++ \cite{zhang2020pointhop++} was proposed to reduce the model complexity of PointHop furthermore. It orders discriminant features by their cross entropy values to improve the performance. The feature learning modules in PointHop and PointHop++ are unsupervised and light-weight. They perform on par with PointNet and surpass unsupervised methods in the point cloud classification task. PointHop++ will be exploited for the design of an UFF encoder in this work. 

{\bf Multi-task Feature Learning.} Multi-task learning exploits commonalities across multiple related tasks so as to complete them simultaneously using the same feature set. They improve efficiency and effectiveness of multiple single-task models. It learns a representation that captures the common information required by multiple tasks. Preliminary research on multi-tasking learning for 2D image-based vision problems. Multi-task learning for 3D point clouds was investigated in \cite{doersch2017multi, hassani2019unsupervised}, where unsupervised tools such as clustering, autoencoding and self-supervised learning are used. 

\begin{figure}[t]
\begin{center}
\includegraphics[width=1\linewidth]{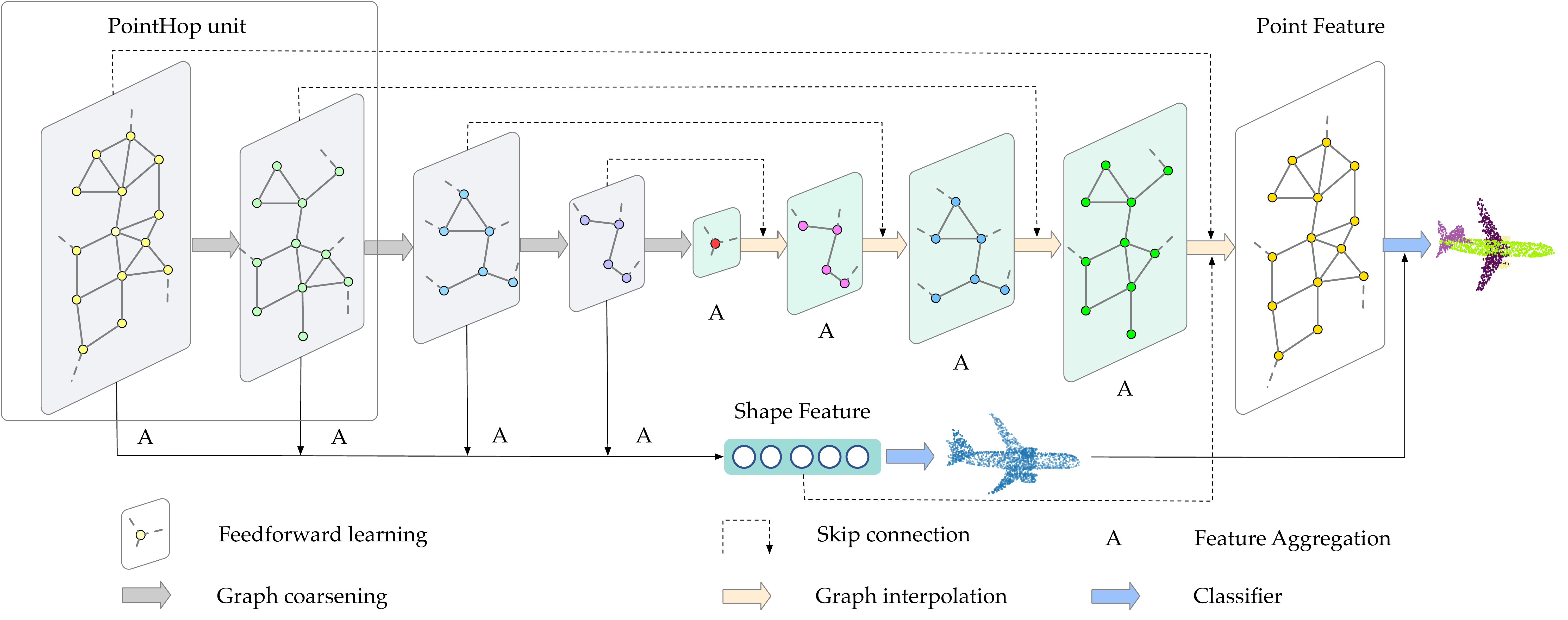}
\end{center}
\caption{An overview of the proposed unsupervised feedforward feature (UFF) learning system, which consists of a fine-to-coarse encoder and a coarse-to-fine decoder.}
\label{fig:fg-1}
\end{figure}

\section{Proposed UFF Method} \label{sec:method}

{\bf System Overview.} We propose an unsupervised feedforward feature (UFF) learning system for joint point cloud classification and segmentation in this section. An illustration of the UFF system is given in Fig. \ref{fig:fg-1}. It takes a point cloud as the input and generates its shape and point features as the output. The UFF system is composed of a fine-to-coarse (F2C) encoder and a coarse-to-fine (C2F) decoder which are in cascade. Such an architecture is frequently used in image segmentation. The encoder provides global shape features while the concatenated encoder-decoder generates local point features. The parameters of the encoder and the decoder are obtained in a feedforward one-pass manner. They are computed using the correlation of spatial coordinates of points in a point cloud set. Since it is a statistics-centric (rather than optimization-centric) approach, no label (or iterative optimization via backpropagation) is needed. 

{\bf Encoder Architecture.} We design the UFF system with the joint classification and segmentation tasks as presented in Sec. \ref{sec:experiment} in mind. The same design principle can be easily generalized to different contexts. The encoder has four layers, where each layer is a PointHop unit \cite{zhang2020pointhop}. A PointHop unit is used to summarize the information of a center pixel and its neighbor pixels. The main operations include: 1) local neighborhood construction using the nearest-neighbor search algorithm, 2) 8-quadrant 3D spatial partitioning and averaging of point attributes from the previous layer in each quadrant for feature extraction, and 3) feedforward convolution using the Saab transform for dimension reduction. A point pooling operation is adopted between two consecutive PointHop units based on the farthest point sampling (FPS) principle. By applying the FPS iteratively, we can reduce the sampled points of a point cloud and enlarge the receptive field from one layer to the next. The PointHop units from the first to the fourth layers summarize the structures of 3D neighborhoods of the short-, mid- and long-range distances, respectively. For more details, we refer to \cite{zhang2020pointhop}. 

{\bf Roles of Encoder/Decoder.} For point-wise segmentation, we need to find discriminant features for all points in the original input point cloud. The down-sampled resolution has to be interpolated back to a finer resolution layer by layer. The spatial coordinates and attributes of points at each layer are recorded by the encoder. The decoder is used to generate new attributes of points layer by layer in a backward fashion. It is important to emphasize the difference between attributes of a point at the encoder and at the decoder. An attribute vector of a point at the encoder is constructed using a bottom-up approach. It does not have a global view in earlier layers. An attribute vector of a point at the decoder is constructed using a bottom-up approach followed by a top-down approach. It has the global information built-in automatically. For convenience, we order layers of the decoder backwards. The inner most layer is the $4^{th}$ layer, then, $3^{rd}$, $2^{nd}$, and $1^{st}$. The outputs of the corresponding layers (with the same scale) between the encoder and the decoder are skip-connected as shown in Fig. \ref{fig:fg-1}. 

{\bf Decoder Architecture.} The decoder is used to obtain features of points at the $(l-1)^{th}$ layer based on point features at the $l^{th}$ layer. Its operations are similar to those of the encoder with minor modification. For every point at the $(l-1)^{th}$ layer, we perform the nearest neighbor search to find its neighbor points located at the $l^{th}$ layer. Then, we conduct the 8-quadrant spatial partitioning and averaging of point attributes at the $l^{th}$ layer in each quadrant for feature extraction. Finally, we perform the feedforward convolution using the Saab transform for dimension reduction after aggregating the features from every quadrants. It is worthwhile to emphasize that the difference between our decoder and that of PointNet++ \cite{qi2017pointnet++}. The latter calculates the weighted sum of the features of the neighbors according to their normalized spatial distances. 

{\bf Feature Aggregation.} Feature aggregation was introduced in \cite{zhang2020pointhop} to reduce the dimension of a feature vector while preserving its representation power. For an $D$-dimensional vector ${\bf a}=(a_1, \cdots, a_D)^T$, $M$ aggregated values can be used to extract its key information, where $M < D$. Then, we can reduce the dimension of the vector from $D$ to $M$. Four ($M=4$) aggregation schemes were adopted in \cite{zhang2020pointhop}. They include the mean, $l_1$-norm, $l_2$-norm and $l_\infty$-norm (i.e., max-pooling) of the input vector. We will apply the same feature aggregation scheme here. Feature aggregation is denoted by $A$ in Fig. \ref{fig:fg-1}. 

Let ${N^l}$ and ${D^l}$ be the point number and the attribute dimension per point at the $l^{th}$ layer, respectively. For the encoder, the raw feature map of a point cloud at the $l^{th}$ layer is a 2D tensor of dimension $N^l \times D^l$. Feature aggregation is conducted along the point dimension, and the aggregated feature map is a 2D tensor of dimension $M \times D^l$, where $M=4$ is the number of aggregation methods. For the decoder, feature aggregation is conducted on points of a finer scale. At the $l^{th}$ layer, after interpolation and point pooling, the raw feature map at each point is a 2D tensor of dimension $S \times D^{l-1}$, where $S=8$ is the number of quadrants. Feature aggregation is conducted along the $S$ dimension, and the aggregated feature map is a 2D tensor of dimension $M \times D^{l-1}$ with $M=4$.

{\bf Integration with Classifiers.} The application of learned shape and point features to point cloud classification and segmentation tasks is also illustrated in Fig. \ref{fig:fg-1}. For point cloud classification, the responses from all layers of the encoder are concatenated and aggregated as shape features. They are then fed into a classifier. No decoder is needed for the classification task. For part segmentation, attributes of a point in the output layer of the decoder are concatenated to get point features. We use the predicted object label to guide the part segmentation task. That is, for each object class, we train a separate classifier for part segmentation. Although feature learning is unsupervised, class and segmentation labels are needed to train classifiers in final decision making. 

\section{Experiments} \label{sec:experiment}

Experiments are conducted to demonstrate the power of UFF features in this section. 

{\bf Model Pre-training.} We obtain UFF model parameters (i.e., the feedforward convolutional filter weights of the Saab transform) from the ShapeNet dataset \cite{chang2015shapenet}. It contains 55 categories of man-made objects (e.g., airplane, bag, car, chair, etc.) and 57,448 CAD models in total. Each CAD model is sampled to 2048 points initially, where each point has three Cartesian coordinates. No data augmentation is used. To show generalizability of the UFF method, we apply the learned UFF model to other datasets without changing the filter weights. This is called the pre-trained model in the following. 

{\bf Shape Classification.} We obtain the shape features from the ModelNet40 dataset \cite{wu20153d} using the pre-trained model. ModelNet40 is a benchmarking dataset for shape classification. It contains 9843 training samples, 2468 testing samples and 40 object classes. All models are initially sampled to 2048 points. We train a random forest (RF) classifier, a linear SVM and a linear least square regressor (only the higher one is reported) on learned features and report the classification accuracy on the ModelNet40 dataset in Table \ref{tab:tab-1}. We compare the performance of our model with respect to unsupervised and supervised methods. As shown in the table, our UFF model achieves {\bf 90.4\%} overall accuracy which surpasses existing unsupervised methods. It is also competitive with state-of-the-art supervised models. 

\begin{table}[htbp]
\caption{Comparison of classification results on ModelNet40.}
\begin{center}
\begin{tabular}{c|c|c}
\hline
& Method & OA (\%) \\ \hline\hline
\multirow{4}*{Supervised$^{\mathrm{a}}$} & PointNet \cite{qi2017pointnet} & 89.2\\
& PointNet++ \cite{qi2017pointnet++} & 90.2\\
& PointCNN \cite{li2018pointcnn} & 92.2\\
& DGCNN \cite{wang2018dynamic} & 92.2\\ \hline
\multirow{2}*{Unsupervised$^{\mathrm{a}}$} & PointHop \cite{zhang2020pointhop} & 89.1 \\
& PointHop++ \cite{zhang2020pointhop++} & 91.1 \\ \hline
\multirow{4}*{Unsupervised$^{\mathrm{b}}$} & FoldingNet \cite{yang2018foldingnet} & 88.9 \\
& PointCapsNet \cite{zhao20193d} & 88.9 \\
& MultiTask \cite{hassani2019unsupervised} & 89.1 \\
& Ours & {\bf 90.4} \\ \hline
\multicolumn{3}{l}{$^{\mathrm{a}}$Learning on the ModelNet40 data.} \\
\multicolumn{3}{l}{$^{\mathrm{b}}$Transfer learning from the ShapeNet on the ModelNet40 data.}
\end{tabular}
\end{center}
\label{tab:tab-1}
\end{table}

{\bf Part Segmentation.} Part segmentation is typically formulated as a point-wise classification task. We need to predict a part category for each point. We conduct experiments on the ShapeNetPart dataset \cite{yi2016scalable}, which is a subset of the ShapeNet core dataset, to evaluate the learned point features. The ShapeNetPart dataset contains 16,881 shapes from 16 object categories. Each object category is annotated with two to six parts, and there are 50 parts in total. Each point cloud is sampled from CAD object models and has 2048 points. The dataset is split into three parts: 12,137 shapes for training, 1,870 shapes for validation and 2,874 shapes for testing. We follow the evaluation metric in \cite{qi2017pointnet}, which is the mean Intersection-over-Union (mIoU) between point-wise ground truth and prediction. We first compute shape's IoU by averaging IoUs of all parts in a shape and, then, obtain mIoUs for a category by averaging over all shapes in the same category. Finally, the instance mIoU (Ins. mIoU) is computed by averaging over all shapes while the category mIoU (Cat. mIoU) is computed by averaging over mIoUs with respect to all categories. 

\begin{table}[htbp]
\caption{Performance comparison on the ShapeNetPart segmentation task with semi-supervised DNNs.}
\begin{center}
\begin{tabular}{c|cccc}
\hline
\multirow{2}*{Method$^{\mathrm{*}}$} & \multicolumn{2}{c}{1\% train data} & \multicolumn{2}{c}{5\% train data} \\ \cline{2-5}
& OA (\%) & mIoU (\%) & OA (\%) & mIoU (\%) \\ \hline\hline
SO-Net \cite{li2018so} & 78.0 & 64.0 & 84.0 & 69.0 \\
PointCapsNet \cite{zhao20193d} & 85.0 & 67.0 & 86.0 & 70.0 \\
MultiTask \cite{hassani2019unsupervised} & 88.6 & 68.2 & 93.7 & 77.7 \\ \hline
Ours & 88.7 & 68.5 & 94.5 & 78.3 \\ \hline
\multicolumn{5}{l}{$^{\mathrm{*}}$Transfer learning from the ShapeNet on the ShapeNetPart data.}
\end{tabular}
\end{center}
\label{tab:tab-2}
\end{table}

\begin{table}[htbp]
\caption{Ablation study of the object-wise segmentation.}
\begin{center}
\newcommand{\tabincell}[2]{\begin{tabular}{@{}#1@{}}#2\end{tabular}}
\begin{tabular}{c|c|c|c}
\hline
Object-wise & Object label & Cat. mIoU (\%) & Ins. mIoU (\%)\\
\hline\hline
No & - & 71.5 & 74.9 \\
Yes & Predicted & {\bf 76.2} & {\bf 78.3} \\
Yes & Ground truth & 78.1 & 81.5 \\
\hline
\end{tabular}
\end{center}
\label{tab:tab-3}
\end{table}

By following \cite{zhao20193d}, we randomly sample 1\% and 5\% of the training data and get point features from sampled training and testing data with the pre-trained model to see both the capability of learning from limited data and the generalization ability on segmentation task. Here we use the predicted labels for the test objects to guide the part segmentation task. Specifically, we first classify the object labels of the test data using the shape feature. Then, we train 16 different random forests on extracted point features of the sampled training data for each object class and evaluate them on the point features of the testing data, which has the corresponding predicted label, respectively. We show the part segmentation results in Table \ref{tab:tab-2} and compare them with three state-of-the-art semi-supervised works. It shows that our method has a better performance. We further conduct an ablation study to validate the object-wise segmentation method in Table \ref{tab:tab-3}. We follow above setting, randomly sample 5\% ShapeNetPart data. It shows that using the predicted test label to guide the part-segmentation increases Ins. mIoU by 3.4\%. Besides, training multiple classifiers reduces the computation complexity. 

\begin{figure}[htbp]
\begin{center}
\includegraphics[width=0.5\linewidth]{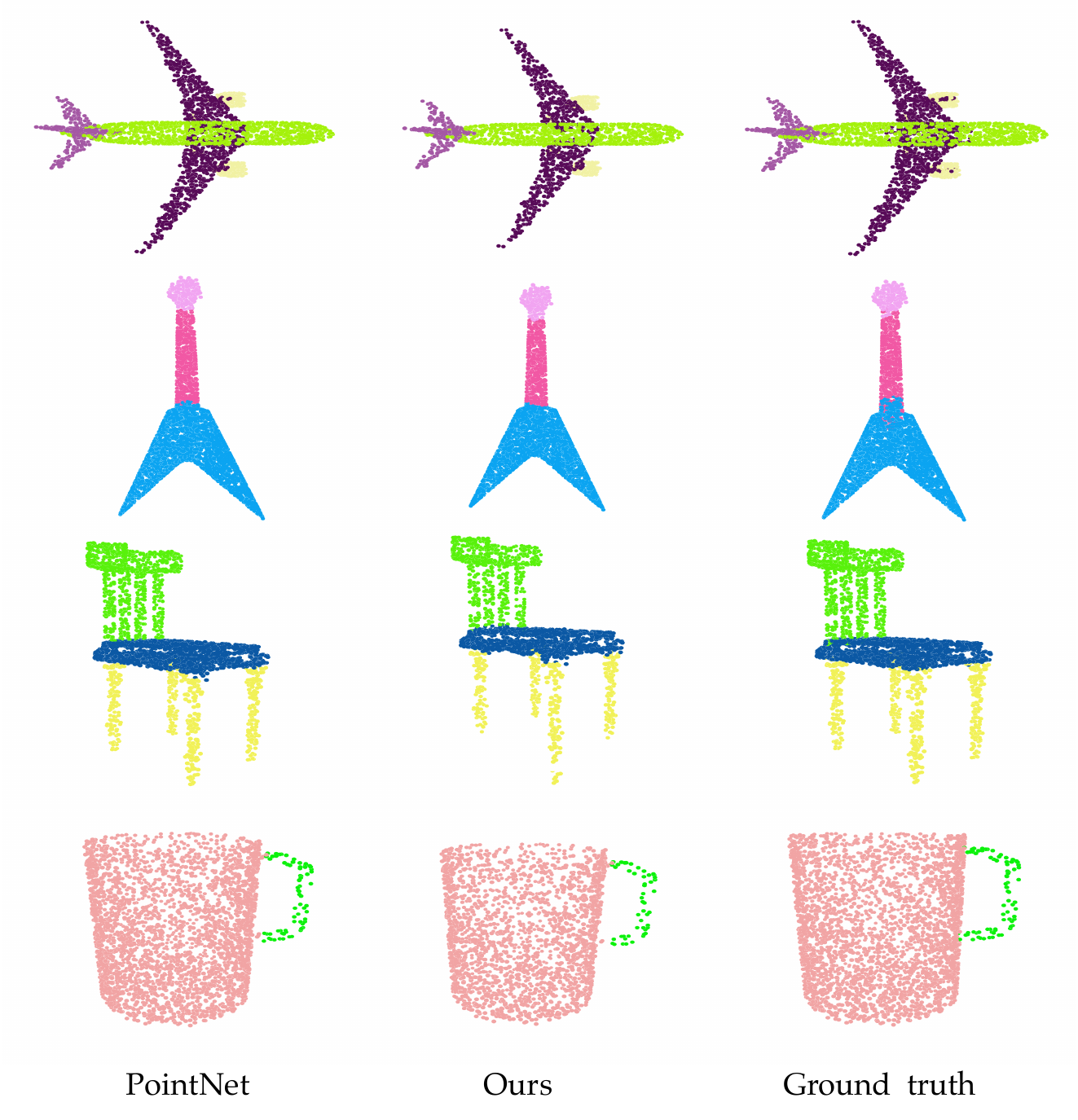}
\end{center}
\caption{Visualization of part segmentation results (from left to
right): PointNet, UFF, the ground truth.}\label{fig:fg-2}
\end{figure}

We compare the UFF method with other state-of-the-art supervised methods in Table \ref{tab:tab-4}. Here, we train the model with 5\% of the ShapeNetPart data (rather than using the pre-trained one). We see a performance gap between our model and DNN-based models. As compared with Table \ref{tab:tab-3}, pre-training boosts the performance. This is consistent with the classification result in Table \ref{tab:tab-1}. Some part segmentation results of PointNet, UFF and the ground truth are visualized in Fig. \ref{fig:fg-2}. In general, visualization results are satisfactory although our model may fail to classify fine-grained details in some hard examples. 

\begin{table}[htbp]
\caption{Performance comparison on the ShapeNetPart segmentation task with unsupervised DNNs.}
\small
\begin{center}
\newcommand{\tabincell}[2]{\begin{tabular}{@{}#1@{}}#2\end{tabular}}
\setlength{\tabcolsep}{0.15mm}{
\begin{tabular}{c|c|c|c|cccccccccccccccc}
\hline
Method$^{\mathrm{*}}$ & \tabincell{c}{\%train \\ data} & \tabincell{c}{Cat. \\ mIoU (\%)} & \tabincell{c}{Ins. \\ mIoU (\%)} & areo & bag & cap & car & chair & \tabincell{c}{ear \\ phone} & guitar & knife & lamp & laptop & motor & mug & pistol & rocket & \tabincell{c}{skate \\ board} & table \\ \hline\hline
PointNet \cite{qi2017pointnet} & \multirow{4}*{100\%} & 80.4 & 83.7 & 83.4 & 78.7 & 82.5 & 74.9 & 89.6 & 73.0 & 91.5 & 85.9 & 80.8 & 95.3 & 65.2 & 93.0 & 81.2 & 57.9 & 72.8 & 80.6 \\
PointNet++ \cite{qi2017pointnet++} & & 81.9 & 85.1 & 82.4 & 79.0 & 87.7 & 77.3 & 90.8 & 71.8 & 91 & 85.9 & 83.7 & 95.3 & 71.6 & 94.1 & 81.3 & 58.7 & 76.4 & 82.6 \\
DGCNN \cite{wang2018dynamic} & & 82.3 & 85.1 & 84.2 & 83.7 & 84.4 & 77.1 & 90.9 & 78.5 & 91.5 & 87.3 & 82.9 & 96.0 & 67.8 & 93.3 & 82.6 & 59.7 & 75.5 & 82.0 \\
PointCNN \cite{li2018pointcnn}& & 84.6 & 86.14 & 84.1 & 86.5 & 86.0 & 80.8 & 90.6 & 79.7 & 92.3 & 88.4 & 85.3 & 96.1 & 77.2 & 95.3 & 84.2 & 64.2 & 80.0 & 83.0\\ \hline
UFF (Ours) & 5\% & 73.9 & 76.9 & 71.9 & 68.8 & 74.9 & 68.0 & 84.4 & 78.2 & 86.2 & 76.1 & 67.7 & 94.5 & 58.0 & 93.2 & 67.5 & 49.9 & 68.0 & 75.6 \\
\hline
\multicolumn{20}{l}{$^{\mathrm{*}}$Learning on the ShapeNetPart data.} \\
\end{tabular}}
\label{tab:tab-4}
\end{center}
\end{table}

\section{Conclusion and Future Work} \label{sec:conclusion}

An UFF method, which exploits statistical correlations of points to learn shape and point features of a point cloud in a one-pass feedforward manner, was proposed in this work. It learns global shape features through an encoder and local point features through the cascade of an encoder and a decoder. Experiments were conducted on joint point cloud classification and part segmentation to demonstrate the power of features learned by UFF. To boost the performance of the shape classification and the part segmentation tasks furthermore, we need to design more powerful classifiers to take full advantages of UFF-learned features. Most classifiers in machine learning are optimized in a single stage. We would like to design more powerful multi-stage classifiers that minimize a predefined cost function as a future extension. 

\bibliographystyle{unsrt}  
\bibliography{main}

\end{document}